\pdfoutput=1

\documentclass[11pt]{article}

\usepackage[final]{acl}

\usepackage{times}
\usepackage{latexsym}

\usepackage[T1]{fontenc}

\usepackage[utf8]{inputenc}

\usepackage{microtype}

\usepackage{inconsolata}

\usepackage{graphicx}
\usepackage{tabularx}
\usepackage{multirow}
\usepackage{algorithm}
\usepackage{algorithmic}
\usepackage{multicol}
\usepackage{lipsum} 
\usepackage{amsmath} 
\usepackage{booktabs}
\hypersetup{
    colorlinks=true,
    linkcolor=black,
    filecolor=green,      
    urlcolor=magenta,
}
\title{Visual Prompting in LLMs for Enhancing Emotion Recognition}

\author{
    Qixuan Zhang\textsuperscript{\rm 1}\footnotemark[1] \hspace{3mm}
    Zhifeng Wang\textsuperscript{\rm 1}\thanks{~~Equal contribution} \hspace{3mm}
    Dylan Zhang\textsuperscript{\rm 2} \hspace{3mm}
    Wenjia Niu\textsuperscript{\rm 3} \\
    \textbf{Sabrina Caldwell}\textsuperscript{\rm 1} \hspace{5mm}
    \textbf{Tom Gedeon}\textsuperscript{\rm 1,4} \hspace{5mm}
    \textbf{Yang Liu}\textsuperscript{\rm 1}\footnotemark[2] \hspace{5mm}
    \textbf{Zhenyue Qin}\textsuperscript{\rm 5}\thanks{~~Corresponding authors} \\
    \textsuperscript{\rm 1}Australian National University \quad 
    \textsuperscript{\rm 2}Quriosity Pty Ltd \\
    \textsuperscript{\rm 3}Webumate Pty Ltd \quad
    \textsuperscript{\rm 4}Curtin University \quad
    \textsuperscript{\rm 5}Yale University \\
    \texttt{\{qixuan.zhang, zhifeng.wang, sabrina.caldwell\}@anu.edu.au} \\
    \texttt{info@quriosity.com.au}, \texttt{niuwenjia9064@hotmail.com} \\
    \texttt{tom.gedeon@curtin.edu.au}, \texttt{\{yang.liu1082, kf.zy.qin\}@gmail.com}\\
Project Page: \url{https://wangzhifengharrison.github.io/sov_prompts_emnlp}
}

\begin{document}

\maketitle
\begin{abstract}
Vision Large Language Models (VLLMs) are transforming the intersection of computer vision and natural language processing. Nonetheless, the potential of using visual prompts for emotion recognition in these models remains largely unexplored and untapped. Traditional methods in VLLMs struggle with spatial localization and often discard valuable global context. To address this problem, we propose a Set-of-Vision prompting (SoV) approach that enhances zero-shot emotion recognition by using spatial information, such as bounding boxes and facial landmarks, to mark targets precisely. SoV improves accuracy in face count and emotion categorization while preserving the enriched image context. Through a battery of experimentation and analysis of recent commercial or open-source VLLMs, we evaluate the SoV model's ability to comprehend facial expressions in natural environments. Our findings demonstrate the effectiveness of integrating spatial visual prompts into VLLMs for improving emotion recognition performance. 
\end{abstract}
 
\section{Introduction}
\label{sec:intro}
As the integration of computer vision and natural language processing progresses, VLLMs \cite{dai2024instructblip} are revolutionizing the way machines interpret visual and textual data. Emotion recognition is gaining considerable interest across multiple disciplines and presents distinct challenges \cite{yang2023context}. It requires the decoding of emotions from nuanced indicators like facial expressions, body language, and contextual details.
\begin{figure}[htbp]
  \centering
  \includegraphics[width = \linewidth]{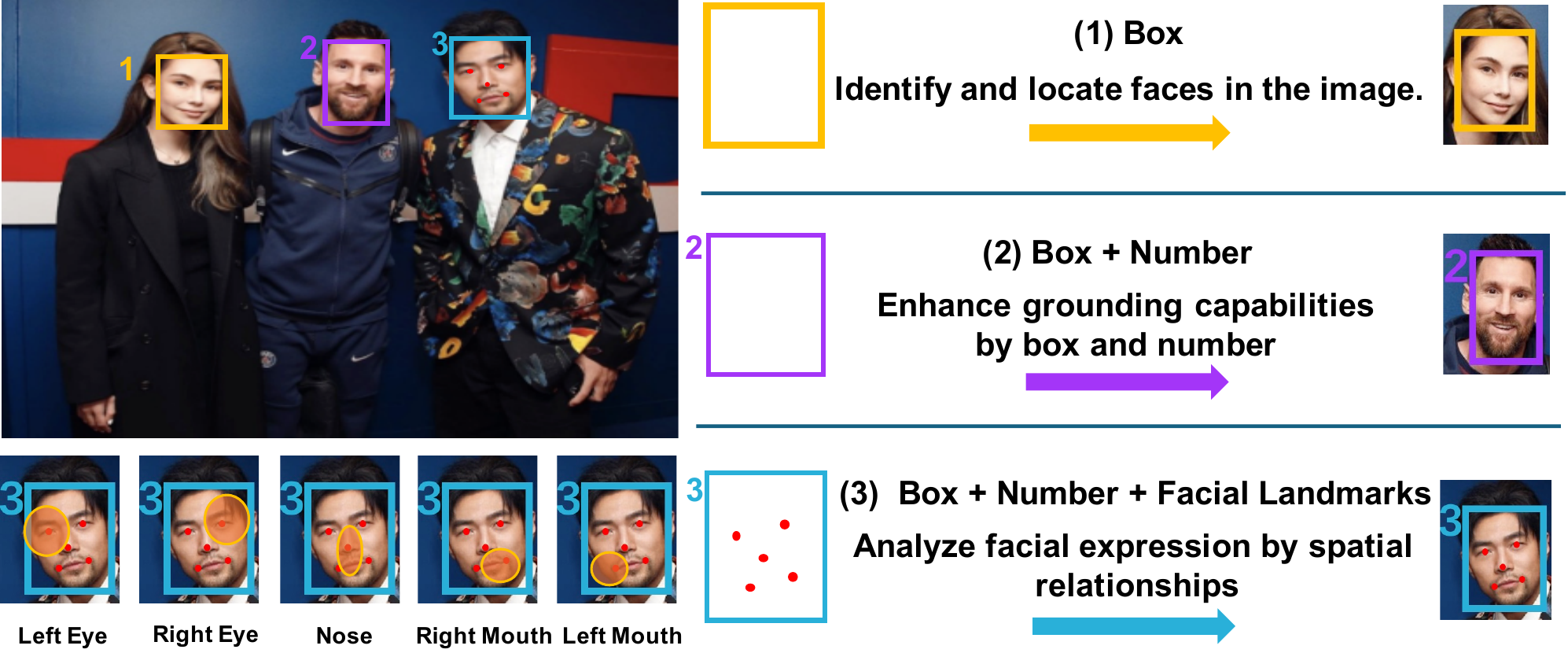}
  \caption{\textbf{Proposed Set-of-Vision (SoV) prompting approach for enhancing facial expression recognition in Vision-Language Large Models (VLLMs)}. SoV progressively incorporates (1) bounding boxes to identify and locate faces, (2) numbered boxes to ground and differentiate faces, and (3) facial landmarks to analyze spatial relationships for fine-grained emotion classification. This multi-stage visual prompting strategy enables VLLMs to accurately detect and recognize emotions in real-world images while preserving global context.
  }
  \label{introduction_four_steps_v2}
\end{figure}

Previous methods, such as those in \cite{xenos2024vllms} and \cite{zhang2023learning}, enhance in-context emotion classification by training transformer-based models or CLIP to generate descriptions of emotions in visual contexts. However, these methods overlook the spatial relationships between different people and facial features within a single face. These spatial relationships can be labeled with numbers and bounding boxes in our SoV prompts to guide VLLMs. A person nearby may have similar facial expressions, while a person farther away may show different facial expressions. Additionally, the relationships between the eyes, mouth, and nose features can be highlighted by facial landmarks in SoV prompts to guide VLLMs.

Recent studies \cite{yang2024fine, zou2024segment,yang2023set, zou2024segment} have explored Visual Prompting, a technique employed in image-language tasks to guide LLMs by incorporating markers such as colorful boxes or circles to emphasize specific targets within an image. 
\begin{figure*}[tb]
  \centering
  \includegraphics[width = \linewidth]{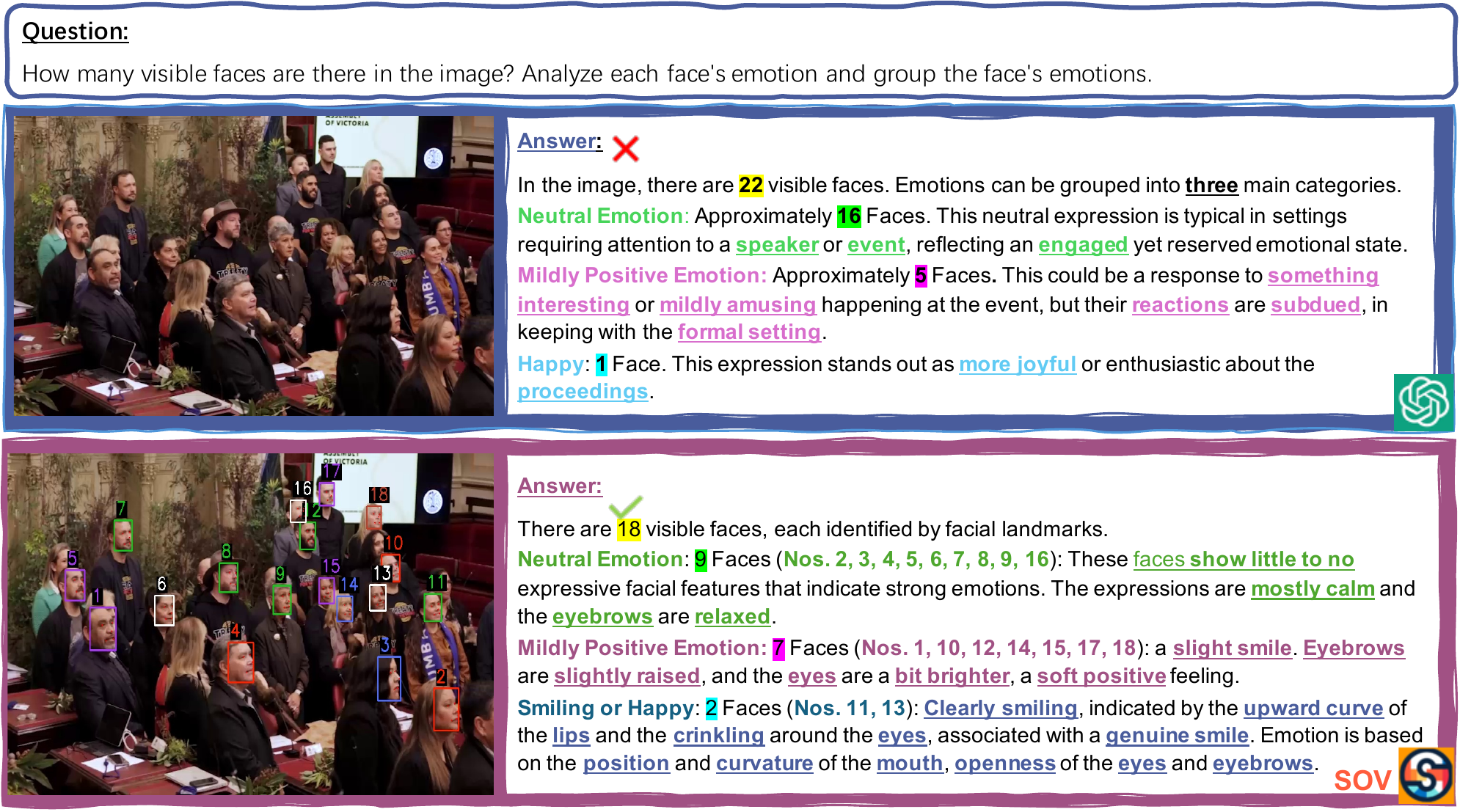}
  \caption{\textbf{Comparative analysis of emotion recognition methods in a group setting}: assessing the precision of facial emotion categorization and face detection using plain text prompts versus Set-of-Vision (SoV) prompts incorporating facial landmarks, bounding boxes, and face enumeration. \textbf{Top}: Results using plain text prompts. \textbf{Bottom}: Results using Set-of-Vision (SoV) prompts. The use of SoV prompts, such as numbering each face, placing bounding boxes, and identifying facial landmarks, allows for a more precise analysis. }
  \label{introduction_motivation}
\end{figure*}
ReCLIP \cite{subramanian2022reclip} adds colorful boxes directly onto an image to highlight specific targets and blurs other irrelevant areas to reduce the performance gap with supervised models on both real and synthetic datasets. Additionally, RedCircle \cite{shtedritski2023does} employs visual prompt engineering, specifically drawing a red circle around an object in an image, to direct a Vision-Language Model's attention to that region and enhance its performance in tasks like zero-shot keypoint localization. However, both of these approaches focus on local objects and ignore spatial context information. Yang \textit{et al.} \cite{yang2024fine} propose using fine-grained visual prompts, such as segmentation masks, and enhancing focus on relevant areas with a `Blur Reverse Mask' that blurs regions outside the target mask to minimize distractions and maintain spatial context. Although visual prompting techniques have garnered interest, their full potential remains unexplored for emotion recognition tasks. Current approaches rely solely on coarse markers like colorful boxes, circles, or masks, which can introduce ambiguity, blur the face images, and pose challenges for accurate recognition tasks. This paper addresses these issues by systematically organizing and investigating various forms of visual prompting. Furthermore, we propose a new prompting approach called Set-of-Vision prompting (SoV) in Fig.~\ref{introduction_four_steps_v2}, which utilizes spatial information such as numbers, bounding boxes, and facial landmarks to precisely mark each target while maintaining background context, thereby enhancing the zero-shot performance of facial expression recognition.
\begin{figure*}[htb]
  \centering
  \includegraphics[width = \linewidth]{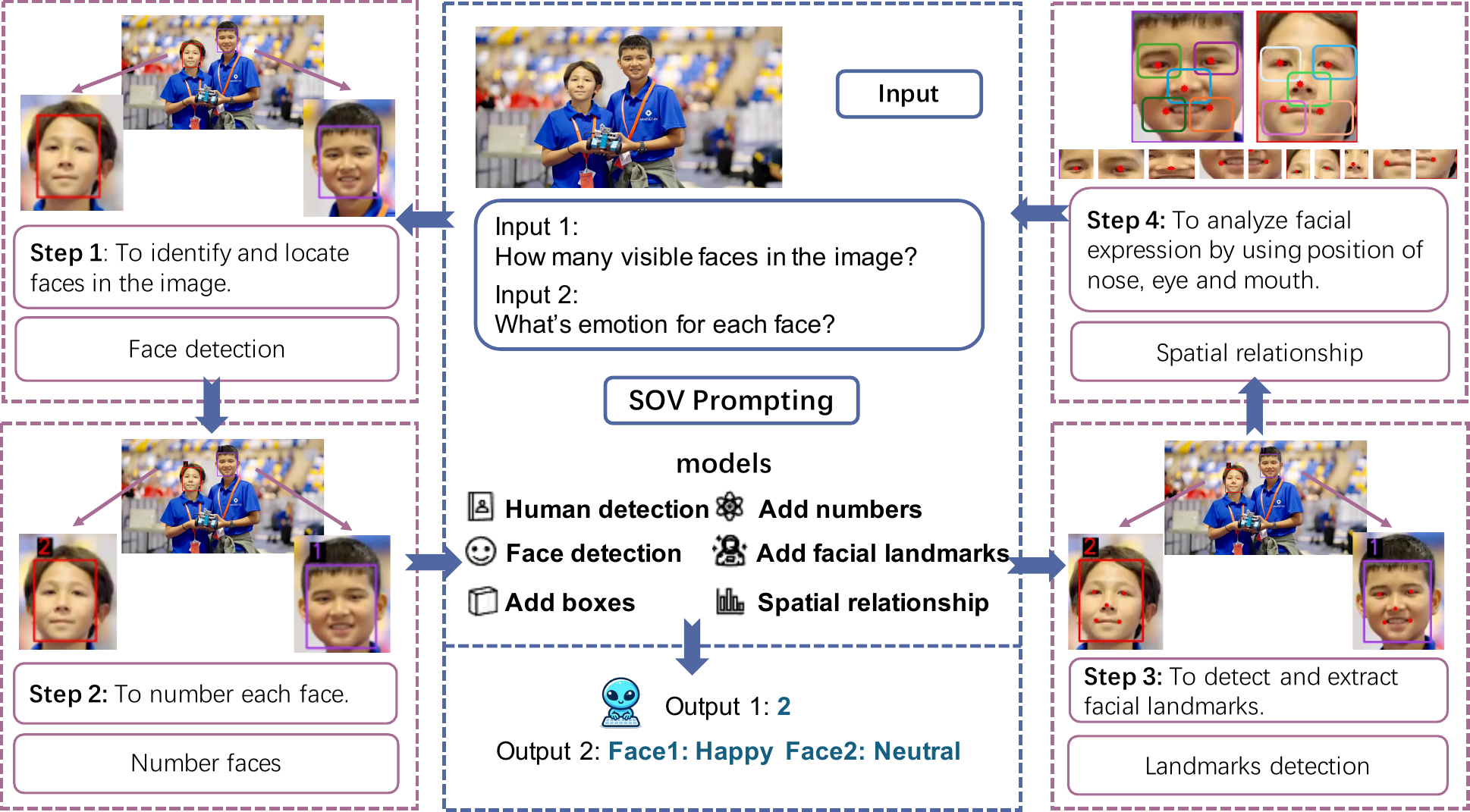}
  \caption{\textbf{Workflow diagram for enhanced face recognition and emotion analysis using the Set-of-Vision (SoV) prompting approach}: a multi-step process involving face detection, face numbering, landmark extraction, and spatial relationship analysis for emotion classification. Each detected face is analyzed and identified by facial landmarks on the face, such as the positions of the nose, eyes, mouth, and other facial features.}
  \label{method_architecture_new}
\end{figure*}
The top of Fig.~\ref{introduction_motivation}, shows an approach where specific vision prompts are not used. As a result, the analysis inaccurately counts 22 visible faces and misclassifies persons' emotions into incorrect categories, with a number of faces labeled under `Neutral Emotion' and fewer under `Mildly Positive Emotion' and `Happy'. This misclassification and miscount demonstrate the limitations when detailed visual cues are not utilized in the analysis. In the bottom of Fig.~\ref{introduction_motivation}, the use of SoV prompts, such as numbering each face, placing bounding boxes, and identifying facial landmarks, allows for a more precise analysis. The correct number of faces is identified (18), and the emotions are accurately categorized into more nuanced groups: `Neutral Emotion', `Mildly Positive Emotion', and `Smiling or Happy'. This method provides a clearer and more detailed breakdown of each individual's emotional state based on visible facial expressions. This comparison highlights the importance and effectiveness of integrating visual prompts in VLLMs analysis for more accurate and detailed recognition and categorization of human emotions in images.
\vspace{-5pt}

To summarize, our main contributions are: (1) The paper introduces a novel visual prompting method (SoV) that highlights facial regions directly within the entire image. This preserves background context, enhancing the ability of VLLMs to perform accurate emotion recognition without the need for cropping faces, thus maintaining the holistic view of the image. (2) The proposed face overlap handling algorithm effectively addresses conflicts arising from overlapping face detections, especially in images with dense face clusters. By prioritizing larger faces and iteratively checking for overlaps, the algorithm ensures that non-occluded faces are retained for subsequent emotion analysis. (3) Our results show that incorporating spatial visual prompts (SoVs) into VLLMs can enhance their performance in recognizing emotions.
%

\section{Related Work}
\subsection{Vision Large Language models}
LLMs such as LLaMA \cite{touvron2023llama}, ChatGPT-3 \cite{brown2020language}, ChatGPT-4 \cite{achiam2023gpt}, and PaLM \cite{chowdhery2023palm} have demonstrated remarkable zero-shot transfer capabilities in natural language processing. Recently, VLLMs, which leverage image-text data pairs from the web, have gained prominence in the computer vision domain. MiniGPT-4 \cite{zhu2023minigpt}, a model that combines a visual encoder with an advanced language model, can enable multi-modal capabilities such as generating detailed image descriptions and designing websites from sketches. Video-LLaVA \cite{zhang2023video} is a multi-modal framework that enhances Large Language Models with the ability to understand and generate responses based on both visual and auditory content in videos. LLaVA \cite{liu2023llava} is a newly developed, end-to-end trained, large multimodal model that combines a vision encoder with a language model, demonstrating promising abilities in multimodal chat. Although VLLMs exhibit remarkable capabilities in vision-based tasks such as image segmentation and object detection, they typically require fine-tuning of the vision and text encoders using existing open vocabulary methods when applied to specific tasks. In contrast, this paper proposes a zero-shot architecture for emotion recognition, overcoming the need for task-specific fine-tuning.

\subsection{Prompting methods}
Prompt engineering is a widely employed technique in the field of NLP \cite{strobelt2022interactive, zhou2024image}. AdbGPT \cite{feng2024prompting} is a novel, lightweight approach that leverages few-shot learning and chain-of-thought reasoning in Large Language Models to automatically reproduce bugs from bug reports, mimicking a developer's problem-solving process without the need for training or hard-coding. Although prompts for large language models have been extensively explored, prompts for vision tasks have received less attention and investigation. Yang \textit{et al.} \cite{yang2024fine} propose using fine-grained visual prompts like segmentation masks and a Blur Reverse Mask strategy to focus on relevant areas. The Image-of-Thought (IoT) prompting method \cite{zhou2024image} enhances Multimodal Large Language Models by guiding them to extract and refine visual rationales step-by-step from images, combining visual and textual insights to improve zero-shot performance on complex visual reasoning tasks. Although these methods have shown promise in tasks like semantic segmentation and object grounding, their performance in emotion recognition has been less effective. This is largely because they tend to analyze individual objects in isolation, overlooking global information and specific facial features, which are crucial for accurately interpreting emotions. To address these issues, the proposed approach directly focuses on the fine-grained facial features present in the entire image, preserving spatial information by utilizing bounding boxes, numbers, and facial landmarks.

\section{Methods}
\subsection{Problem Definition}
The task of matching images to emotions for each visible face in a given image involves several sophisticated steps, combining face detection and emotion recognition. Typically, the VLLMs, denoted as $\Phi$, will take an image $I \in \mathcal{R}^{H \times W \times 3}$ and a text question of length $l_i$, $Q^{i}=[q^{i}_{1},q^{i}_{2},..., q^{i}_{l_i}]$, as input. The output is a sequence of answers with length $l_{o}$, containing emotions, $A^{o}=[a^{o}_{1},a^{o}_{2},..., a^{o}_{l_o}]$, which can be formulated as (Eq. \ref{problems_definition}):
\begin{equation}
A^{o}= \Phi(I,Q^{i})
\label{problems_definition}
\end{equation}

\begin{figure}[tbp]
  \centering
  \includegraphics[width = \linewidth]{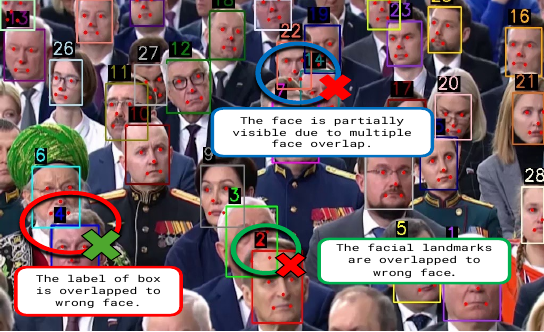}
  \caption{\textbf{Face detection inevitably introduces some overlaps or conflicts that confuse VLLMs.} Analyzing the impact of face overlaps, occlusions, landmark misalignment, and bounding box conflicts for emotion recognition.}
  \label{overlap_images_for_algorithm}
\end{figure}
In our task, we aim to find the best matching image-emotion pairs ($I$, $A^o$) for each visible face. Traditionally, this involves cropping the face from the image using face detectors. However, with the introduction of visual prompting, faces can be directly marked on the entire image, highlighting the facial region while preserving the background context and avoiding the obscuration of faces. With this in mind, we have developed Set-of-Vision prompts (SoV), a simple method of overlaying a number of visual prompts on the facial regions in an image. 
\begin{algorithm}[H]
\caption{Detect and Handle Overlaps}
\begin{small}
\begin{algorithmic}[1]
\STATE // Define a function to check if overlap
\STATE \textbf{def} Boxes\_overlap($B_1, B_2$):    
\STATE \quad $\tilde{O} = Check\_Overlap(B_1, B_2)$ // Check if overlap
\STATE \quad \textbf{return} $\tilde{O}$
\STATE // Define a function to calculate the area
\STATE \textbf{def} Face\_size(B):    
\STATE \quad $\tilde{A} = Calculate\_Area(B)$ // Calculate the area
\STATE \quad \textbf{return} $\tilde{A}$
\STATE // Main function 
\STATE \textbf{def} Detect\_and\_handle\_overlaps(faces):   
\STATE \quad // Sort faces by the area in descending order
\STATE \quad $\tilde{F} = \textbf{Sorted}(faces)$
\STATE \quad // Initialize an empty list 
\STATE
\quad visible\_faces = []
\STATE
\quad \textbf{for} {$k$ in range(K)}    
\STATE \quad \quad \textbf{if} Boxes\_overlap($\tilde{F}[k], \tilde{F}[k-1]$):
\STATE \quad \quad \quad \textbf{if} Face\_size($\tilde{F}[k]$) $>$ Face\_size($\tilde{F}[k-1]$):
\STATE \quad \quad \quad \quad // Replace the smaller face 
\STATE \quad \quad \quad \quad visible\_faces $= \neg \tilde{F}[k-1] \land \tilde{F}[k]$
\STATE \quad \textbf{return} visible\_faces
\end{algorithmic}
\label{overlap_solution}
\end{small}
\end{algorithm}

This operation augments the input image $I$ to a new image $I^{new} = SoV(I)$, while keeping the text prompts to VLLMs unchanged as shown in Fig. \ref{introduction_motivation}. It can be formulated as (Eq. \ref{problems_definition_new}):
\begin{equation}
A^{o}= \Phi(SoV(I),Q^{i})
\label{problems_definition_new}
\end{equation}

\subsection{Set of Vision Prompts}
\subsubsection{Box detection}
Once the image is obtained, we need to generate visual prompts for the image that will be utilized by VLLMs for emotion recognition. We employ the RetinaFace \cite{deng2020retinaface} algorithm to detect faces within the image. Let $B = \{b_1, b_2, \ldots, b_n\}$ denote the set of detected face bounding boxes, where $b_i$ represents the $i$-th face bounding box. The process can be formulated as (Eq. (\ref{box_detection})):
\begin{equation}
b_{i}= \mathcal{D}(I,\theta_{i})
\label{box_detection}
\end{equation}
, where $I$ is the input image; $\theta_{i}$ represents the hyperparameters for the RetinaFace model $\mathcal{D}$; and $b_{i}$ corresponds to the $i$-th face bounding box.
\subsubsection{Box Overlap Handling Algorithm}
However, this face detection algorithm inevitably introduces some overlaps or conflicts that confuse VLLMs, especially in images with densely populated faces, such as when two faces overlap in one area or one face is obscured by another. This is illustrated in Fig. \ref{overlap_images_for_algorithm}. To mitigate this problem, we propose a face overlap handling algorithm, as shown in Algorithm. \ref{overlap_solution}. Given the set of boxes \( B = \{b_1, b_2, \ldots, b_n\} \), we first calculate the area for each bounding box \( b_i \), then sort the detected faces  \(b_i\) by their area in descending order (line 12) (Eq. (\ref{sorted_box})):
\begin{equation}
\centering
\label{sorted_box}
B_{sorted} = \{b_1, b_2, \ldots, b_n\} 
\end{equation}

, where$\quad \text{Area}(b_1) \geq \text{Area}(b_2) \geq \ldots \geq \text{Area}(b_n)$, ensuring that larger faces are prioritized. 

By iterating through the sorted faces, the algorithm checks for overlaps and compares the areas of overlapping faces. For each face \( \tilde{F}[k] \) in \( B_{sorted} \), it checks if \( \tilde{F}[k] \) overlaps with any face \( \tilde{F}[j] \) in \( F_{final} \) by (Eq. (\ref{if_overlap_box})):
\begin{equation}
\begin{split}
       &\text{Overlap}(\tilde{F}[k], \tilde{F}[j]) = \\
       &\frac{\text{Area}(\tilde{F}[k] \cap \tilde{F}[j])}{\min(\text{Area}(\tilde{F}[k]), \text{Area}(\tilde{F}[j]))} > \epsilon
\end{split}
    \label{if_overlap_box}
\end{equation}

If the overlap is significant, compare their areas and discard the smaller face by (Eq. (\ref{discard_boxes})):

\begin{small}
\begin{equation}
\tilde{F}[k] = \begin{cases} 
\hat{F}[k] & \text{if } \text{Area}(\hat{F}[k]) > \text{Area}(\hat{F}[j]), \\
\hat{F}[j] & \text{otherwise}.
\end{cases}
 \label{discard_boxes}
\end{equation}
\end{small}
Add non-occluded faces to \( F_{final} \) by (Eq. (\ref{add_final_list})):
\begin{equation}
       F_{final} \leftarrow F_{final} \cup \{\tilde{F}[k]\}
    \label{add_final_list}
\end{equation}
It ensures that only the faces that are close to the camera and not obstructed remain in the final list. 

Once we determine the location of boxes $B = \{b_1, b_2, \ldots, b_n\}$ for each face, We need to assign a unique ID to each face. The unique ID $N = \{1, 2, \ldots, n\}$ will be used to locate each face in the image, where \( n \) is the number of detected faces. Thus, the set of vision prompts becomes distinguishable and can be effectively interpreted by VLLMs.

\subsubsection{Facial Landmarks Detection and Analysis}
After identifying and handling overlapping faces, we proceed with the detection and extraction of facial landmarks for each face. For each face \( b_i \in F_{final} \):
\begin{equation}
\begin{split}
        L_i &= \Theta(b_i) \\
         &= \{(x_1, y_1), (x_2, y_2), \ldots, (x_m, y_m)\}
\end{split}
\label{landmarks_model}
\end{equation}
, where $\Theta$ is landmarks extraction model, $L_i$ is the facial landmarks, $(x_m, y_m)$ is coordinates of facial landmarks. For each set of landmarks \( L_i \), analyze spatial relationships for facial expression within RGB image:
\begin{equation}
  E(L_i) = f(R(L_i))
\label{spatial_relationship}
\end{equation}
, where \( f \) is a function mapping spatial relationships $R(L_i)$ to facial expressions $ E(L_i)$. The entire process is illustrated in Fig. \ref{method_architecture_new}.
\begin{figure*}[htbp]
  \centering
  \includegraphics[width = \linewidth]{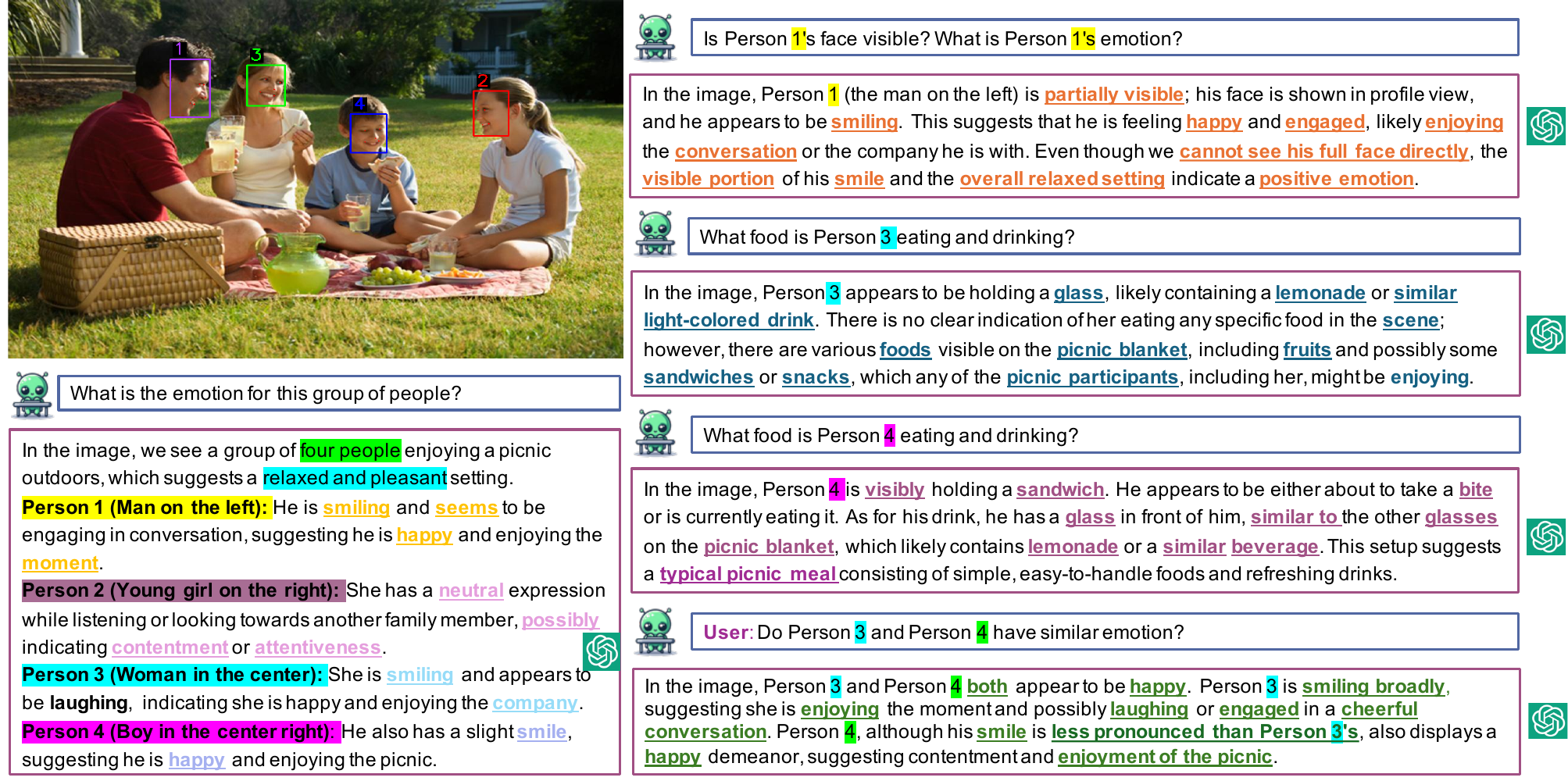}
  \caption{We use two types of prompt methods. \textbf{Left}: plain text prompts, which can be used for group emotion recognition. \textbf{Right}: combined text-vision prompts, which can be used for analyzing specific individuals' emotions. These prompts can be used to evaluate emotional interpretation in social interactions based on facial expressions, body language, and contextual cues.}
  \label{insert_number_image}
\end{figure*}
\subsection{Text and Vision Prompts}
We have a collection of $n$ pairs of location-vision prompts, represented as ${(l_1, v_1),\ldots, (l_n, v_n)}$. When introducing additional text prompts for a new image $I_{new}$, we can choose to use either plain text prompts or a combination of text and vision prompts.
\subsubsection{Plain Text Prompts}
This method is exemplified on the left side in Fig. \ref{insert_number_image}. It involves asking a general question about the emotional state of a group of people without referencing specific individuals. For example, the question "What is the emotion for this group of people?" yields an answer that considers the overall mood and setting of the group. This approach is useful for understanding group dynamics or the general atmosphere of a scene.

\subsubsection{Combined Text-Vision Prompts}
Shown on the right side in Fig.~\ref{insert_number_image}, this method involves more detailed prompts that focus on individual persons within the group. This allows for a more nuanced analysis of specific people's emotions and actions. For instance, questions such as `What is Person 1's emotion?' or `What food is Person 3 eating and drinking?' prompt answers that delve into specific details regarding individuals' facial expressions, body language, and interactions with objects, like food and drinks.

\begin{table*}[ht!]
\centering
\small
\caption{\textbf{Comparison of zero-shot emotion recognition methods}, including MiniGPT-4 \cite{zhu2023minigpt}, LLaVA \cite{liu2023llava}, Video-LLaVA \cite{zhang2023video}, GPT-4V  \cite{achiam2023gpt}, and SoV-Enhanced GPT Models, across datasets with varying difficulty levels (Easy, Medium, and Hard): A Comparative Analysis of Accuracy and Top-1 Recall (R@1).}
\resizebox{\linewidth}{!}{%
\begin{tabular}{c c c c c c c c c c c c}
\toprule
\multirow{2}{*}{\textbf{Methods}} & \multirow{2}{*}{\textbf{Backbone}} & \multicolumn{2}{c}{\textbf{Easy}} & \multicolumn{2}{c}{\textbf{Medium}} & \multicolumn{2}{c}{\textbf{Hard}} & \multicolumn{2}{c}{\textbf{Total}} \\ 
\cmidrule(lr){3-10}
& & \textbf{Acc (\%)} & \textbf{R@1} & \textbf{Acc (\%)} & \textbf{R@1} & \textbf{Acc (\%)} & \textbf{R@1} & \textbf{Acc (\%)} & \textbf{R@1} \\ 
\midrule
\textbf{MiniGPT-4 \cite{zhu2023minigpt}} & Q-former, ViT & 30.45 & 16.17 & 19.88 & 12.85 & 15.78 & 14.10 & 22.87 & 12.96 \\ 
\textbf{LLaVA \cite{liu2023llava}} & CLIP, ViT & 35.74 & 15.91 & 22.80 & 11.29 & 3.50 & 1.58 & 22.65 & 10.56 \\ 
\textbf{Video-LLaVA \cite{zhang2023video}} & Pre-align ViT & 20.11 & 9.37 & 16.95 & 7.26 & 8.77 & 4.46 & 16.12 & 6.84 \\ 
\textbf{GPT-4V \cite{achiam2023gpt}} & ViT & 48.85 & 27.94 & 47.95 & 19.23 & 32.45 & 11.36 & 44.44 & 22.11 \\ 
\midrule
\textbf{GPT-4o \cite{gpt4o} +SoV (Ours)} & ViT & \textbf{\textcolor{blue}{51.27}} & \textbf{\textcolor{blue}{31.93}} & \textbf{\textcolor{blue}{49.12}} & \textbf{\textcolor{blue}{22.65}} & \textbf{\textcolor{blue}{49.12}} & \textbf{\textcolor{red}{20.46}} & \textbf{\textcolor{blue}{50.10}} & \textbf{\textcolor{blue}{24.20}} \\ 
\textbf{GPT-4V \cite{achiam2023gpt} +SoV (Ours)} & ViT & \textbf{\textcolor{red}{60.91}} & \textbf{\textcolor{red}{41.96}} & \textbf{\textcolor{red}{53.21}} & \textbf{\textcolor{red}{22.82}} & \textbf{\textcolor{red}{50.00}} & \textbf{18.97} & \textbf{\textcolor{red}{55.33}} & \textbf{\textcolor{red}{28.69}} \\ 
\bottomrule
\end{tabular}
}
\label{sota_result_compare}
\end{table*}

\section{Experiments}
\subsection{Models and Settings}
We do not need to train any models for our method. We evaluate the model's performance in a zero-shot manner using VLLMs. We include both commercial models such as GPT-4V-turbo \cite{achiam2023gpt}\footnote{https://chatgpt.com/} and GPT-4o-2024-05-13 \cite{gpt4o} as well as open-sourced models including MiniGPT-4-Vicuna \cite{zhu2023minigpt}\footnote{https://github.com/Vision-CAIR/MiniGPT-4},  LLaVA-1.5-7B \cite{liu2023llava}\footnote{https://github.com/haotian-liu/LLaVA}, Video-LLaVA-7B \cite{zhang2023video}\footnote{https://github.com/PKU-YuanGroup/Video-LLaVA}. 

\begin{figure*}[tb]
  \centering
  \includegraphics[width = \textwidth]{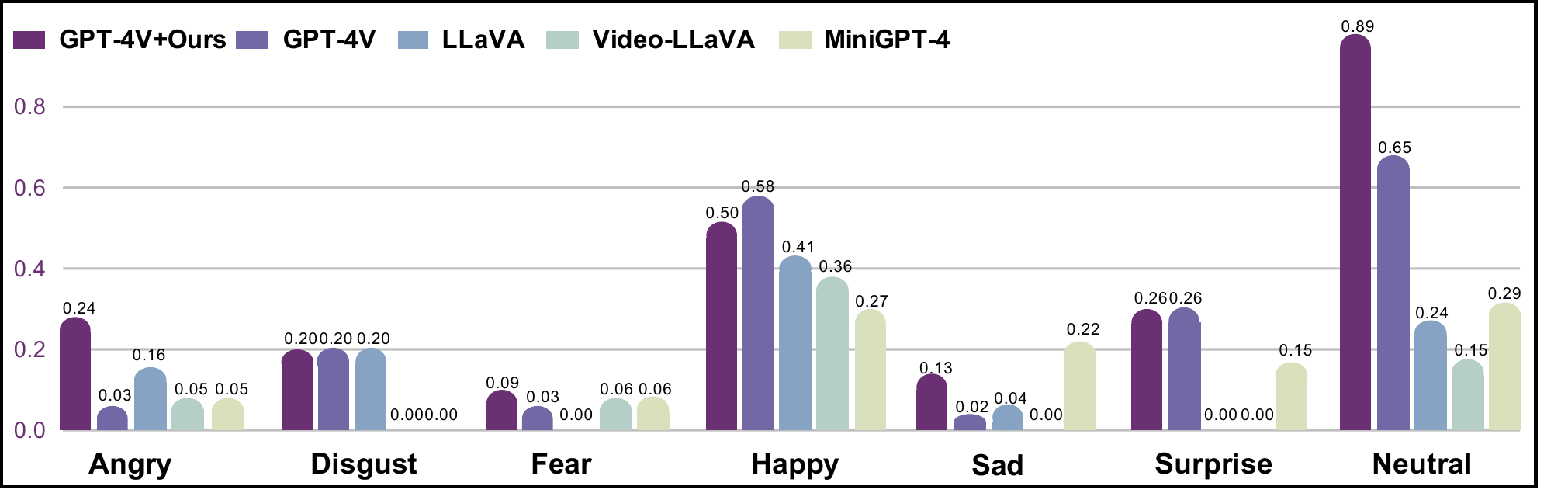}
  \caption{The bar chart shows the performance of various VLLMs in recognizing different emotions from images. The models compared include GPT-4V+Ours, GPT-4V \cite{achiam2023gpt}, LLaVA \cite{liu2023llava}, Video-LLaVA \cite{zhang2023video}, and MiniGPT-4 \cite{zhu2023minigpt}. These results are distributed across seven emotions.
  }
  \label{sota_7_emotions}
\end{figure*}
\subsection{Dataset details} We collect original images from ABC News website\footnote{https://www.abc.net.au/news/}. Following the collection, we undertake meticulous preprocessing, initially removing any identical and blurry images through deduplication. To minimize human effort and cost in data annotation, we employ DeepFace \cite{serengil2024benchmark} for emotion annotation. Subsequently, two human annotators revise and refine the image labels. Finally, to finalize the labels, we involved a third annotator who has a professional background in psychology to verify correctness of facial expressions with their domain knowledge. This procedure guarantees the quality of the annotated data used to construct benchmarks. Table \ref{dataset_introduction} in appendix \ref{appendix_dataset_details} presents the dataset details used for testing a model on the task of zero-shot emotion recognition, structured across three different levels of difficulty: Easy, Medium, and Hard. This structured dataset aids in understanding the robustness and adaptability of the model in varying conditions of visual complexity.

\subsection{Quantitative Results}
Table \ref{sota_result_compare} provides a detailed analysis of different zero-shot emotion recognition methods. MiniGPT-4 \cite{zhu2023minigpt} exhibits low performance, with accuracy ranging from 15.78\% to 30.45\% and Recall@1 from 12.85\% to 16.17\%. LLaVA \cite{liu2023llava} and Video-LLaVA \cite{zhang2023video} perform better in simpler categories but struggle significantly in the Hard category, where accuracy plummets to 3.50\% and Recall@1 to 1.58\%. In contrast, GPT-4V \cite{achiam2023gpt} demonstrates robust performance across all levels, markedly improved by the SoV prompts. Specifically, GPT-4V+SoV achieves an impressive 60.91\% accuracy and 41.96\% Recall@1 in the Easy category, maintaining 50.00\% accuracy and 18.97\% Recall@1 even in the Hard category. These results underline SoV's effectiveness in boosting the model’s ability to accurately interpret emotions across different complexities.
\begin{table*}[ht!]
\centering
\small
\caption{\textbf{Comparison of SOTA methods for zero-shot emotion recognition across datasets with varying levels of difficulty—Easy, Medium, and Hard.} The types of visual prompts used by previous approaches are: \textbf{P}: Crop, \textbf{B}: Box, \textbf{R}: Blur Reverse, \textbf{C}: Circle, \textbf{N}: Number, \textbf{F}: Facial Landmarks.}
\resizebox{\linewidth}{!}{%
\begin{tabular}{c c c c c c c c c c}
\toprule
\multirow{2}{*}{\textbf{SOTA methods}} & \multirow{2}{*}{\textbf{Visual Prompt}} & \multicolumn{2}{c}{\textbf{Easy}} & \multicolumn{2}{c}{\textbf{Medium}} & \multicolumn{2}{c}{\textbf{Hard}} & \multicolumn{2}{c}{\textbf{Total}} \\ 
\cmidrule(lr){3-10}
& & \textbf{Acc (\%)} & \textbf{R@1} & \textbf{Acc (\%)} & \textbf{R@1} & \textbf{Acc (\%)} & \textbf{R@1} & \textbf{Acc (\%)} & \textbf{R@1} \\ 
\midrule
\textbf{Baseline \cite{achiam2023gpt}} & Plain Text & 48.85 & 27.94 & 47.95 & 19.23 & 32.45 & 11.36 & 44.44 & 22.11 \\ 
\textbf{ReCLIP \cite{subramanian2022reclip}} & P | B | R & \textbf{\textcolor{blue}{54.02}} & \textbf{\textcolor{blue}{31.47}} & 46.19 & 16.98 & 42.10 & 14.32 & 48.14 & 22.98 \\ 
\textbf{RedCircle \cite{shtedritski2023does}} & P | C | R & 51.72 & 29.55 & \textbf{\textcolor{blue}{48.53}} & \textbf{\textcolor{red}{23.19}} & \textbf{\textcolor{blue}{45.61}} & \textbf{\textcolor{blue}{15.89}} & \textbf{\textcolor{blue}{49.01}} & \textbf{\textcolor{blue}{23.89}} \\ 
\midrule
\textbf{SoV (Ours)} & N | B | F & \textbf{\textcolor{red}{60.91}} & \textbf{\textcolor{red}{41.96}} & \textbf{\textcolor{red}{53.21}} & \textbf{\textcolor{blue}{22.82}} & \textbf{\textcolor{red}{50.00}} & \textbf{\textcolor{red}{18.97}} & \textbf{\textcolor{red}{55.33}} & \textbf{\textcolor{red}{28.69}} \\ 
\bottomrule
\end{tabular}
}
\label{sota_redclip_red_circle}
\end{table*}
\begin{figure*}[htb]
  \centering
  \includegraphics[width = \linewidth]{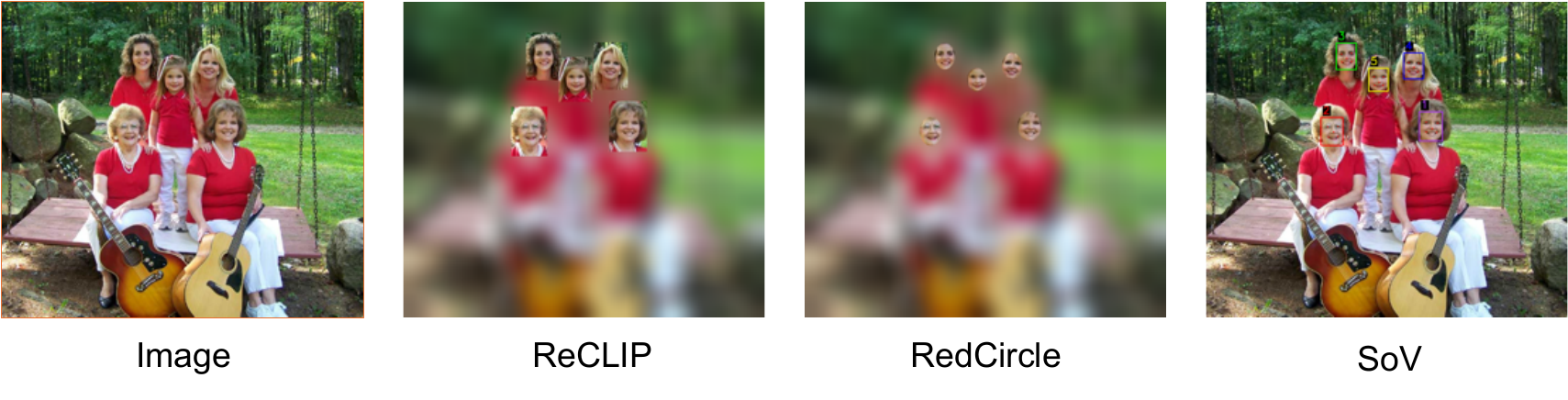}
  \caption{Visualization of the SOTA visual prompting approaches such as ReCLIP \cite{subramanian2022reclip}, RedCircle \cite{shtedritski2023does} and our SoV prompts. ReCLIP and RedCircle blur out non-facial areas and highlight faces with rectangles and circles, respectively. SoV(Ours) applies a combination of visual prompts to emphasize facial areas while maintaining background context.
  }
  \label{four_method_images}
\end{figure*}

In Fig.~\ref{sota_7_emotions}, the chart highlights the varied efficacy of different VLLMs in emotion recognition tasks. GPT-4V+Ours consistently outperforms other models across nearly all emotions, particularly excelling in neutral and angry emotions. This highlights the specialized capabilities of GPT-4V+Ours in capturing more nuanced and varied emotional states. Meanwhile, other models show selective strengths and general weaknesses, particularly in recognizing negative emotions like fear and disgust.
\subsection{Visual Prompting}
Table.~\ref{sota_redclip_red_circle} presents a comparative analysis of state-of-the-art methods for zero-shot emotion recognition across datasets categorized by varying levels of difficulty: Easy, Medium, and Hard. It details performance metrics such as Accuracy and Recall, comparing the effectiveness of different visual prompting strategies utilized by GPT-4V. The baseline method (GPT-4V), which uses plain text prompts, demonstrates moderate effectiveness, with an overall accuracy of 44.44\% and a Recall of 22.11\%. 
\begin{table*}[tb]
\centering
\small
\caption{\textbf{Ablation study for vision prompts on GPT-4V.} \textbf{Baseline}: represents the model's performance without any additional prompts. \textbf{Box}: indicates a visual prompt that uses bounding boxes. \textbf{Box+Number}: adding numerical identifiers to the bounding boxes. \textbf{SoV}: adding facial landmarks to each face with additional numerical identifiers to the bounding boxes.}
\resizebox{\linewidth}{!}{
\begin{tabular}{c c c c c c c c c}
\toprule
\multirow{2}{*}{\textbf{Vision Prompt}} & \multicolumn{2}{c}{\textbf{Easy}} & \multicolumn{2}{c}{\textbf{Medium}} & \multicolumn{2}{c}{\textbf{Hard}} & \multicolumn{2}{c}{\textbf{Total}} \\ 
\cmidrule(lr){2-9}
 & \textbf{Acc (\%)} & \textbf{R@1} & \textbf{Acc (\%)} & \textbf{R@1} & \textbf{Acc (\%)} & \textbf{R@1} & \textbf{Acc (\%)} & \textbf{R@1} \\ 
\midrule
\textbf{Baseline \cite{achiam2023gpt}} & 48.85 & 27.94 & 47.95 & 19.23 & 32.45 & 11.36 & 44.44 & 22.11 \\ 
\textbf{Box} & 47.12 & 29.47 & 45.61 & 17.73 & 39.47 & 12.46 & 44.66 & 23.52 \\ 
\textbf{Box+Number} & \textbf{\textcolor{blue}{58.04}} & \textbf{\textcolor{blue}{41.10}} & \textbf{\textcolor{blue}{51.46}} & \textbf{\textcolor{blue}{22.12}} & \textbf{\textcolor{blue}{42.10}} & \textbf{\textcolor{blue}{15.57}} & \textbf{\textcolor{blue}{51.63}} & \textbf{\textcolor{blue}{28.24}} \\ 
\midrule
\textbf{SoV} & \textbf{\textcolor{red}{60.91}} & \textbf{\textcolor{red}{41.96}} & \textbf{\textcolor{red}{53.21}} & \textbf{\textcolor{red}{22.82}} & \textbf{\textcolor{red}{50.00}} & \textbf{\textcolor{red}{18.97}} & \textbf{\textcolor{red}{55.33}} & \textbf{\textcolor{red}{28.69}} \\ 
\bottomrule
\end{tabular}
}
\label{ablation_table_vision_prompts}
\end{table*}
\begin{figure}[htb]
  \centering
  \includegraphics[width = \linewidth]{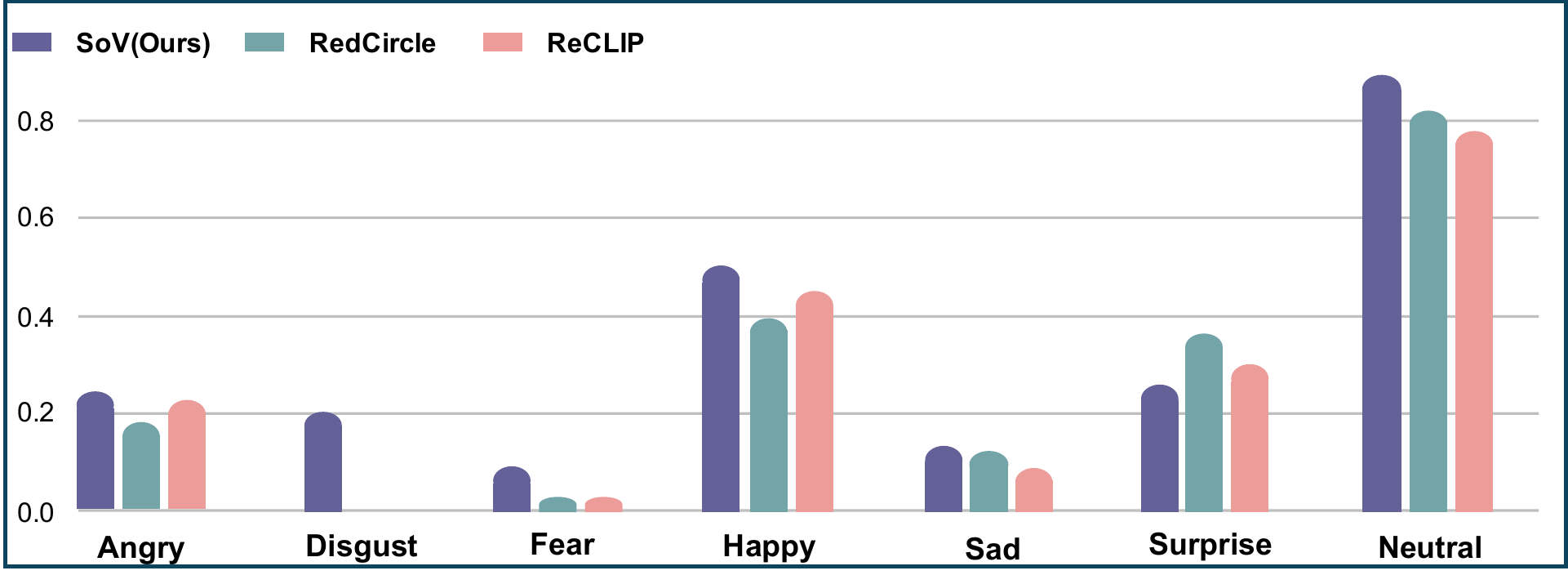}
  \caption{The bar chart illustrates the performance of SoV(Ours), RedCircle and ReCLIP in emotion recognition across seven different emotional categories.
  }
  \label{total_3_methods}
\end{figure}
Methods such as ReCLIP, RedCircle, and SoV employ more complex combinations of visual prompts. SoV (Ours), incorporating Numbers, Boxes, and Facial Landmarks, achieves the highest overall accuracy and recall scores of 55.33\% and 28.69\%, respectively. This suggests that the integration of multiple visual cues, particularly those that enhance the recognition of facial features, significantly improves performance across all difficulty levels, especially in more challenging datasets.

Fig.~\ref{four_method_images} visually represents how each visual prompting approach modifies the image to focus on emotion-relevant features. ReCLIP and RedCircle blur out non-facial areas and highlight faces with rectangles and circles, respectively. SoV applies a combination of visual prompts to emphasize facial areas while maintaining background context, which is critical for emotion recognition.
\begin{figure}[tb]
  \centering
  \includegraphics[width = \linewidth]{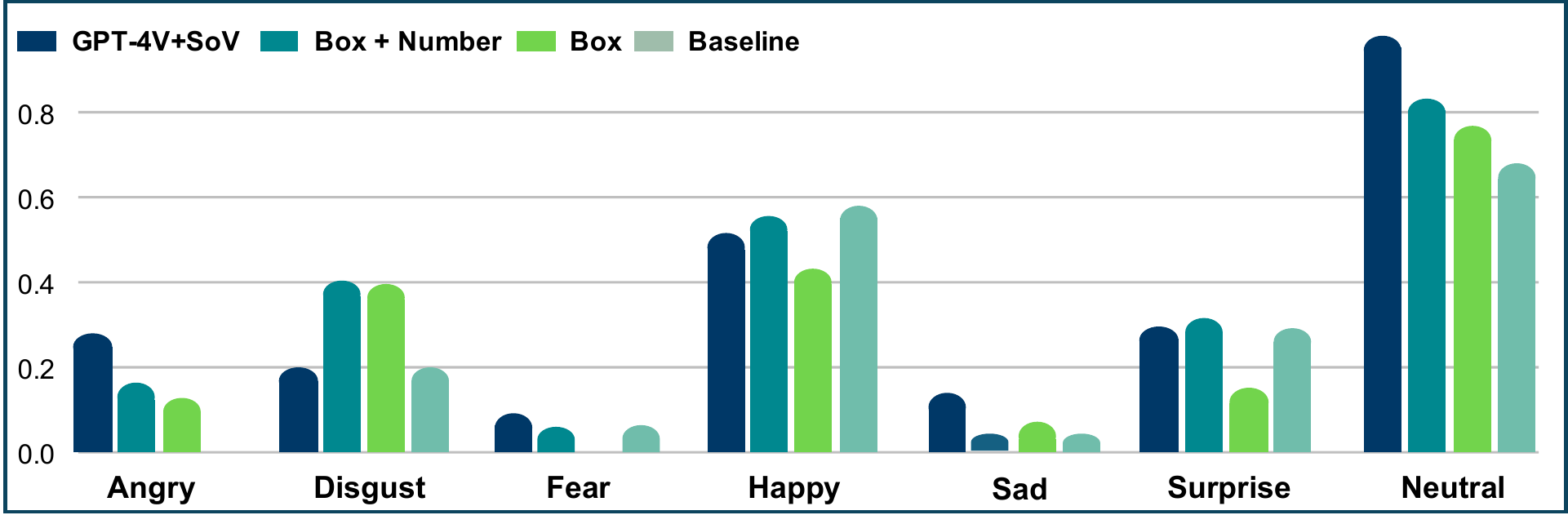}
  \caption{The bar chart displayed in the image illustrates the performance of different vision prompts—GPT-4V+SoV, Box + Number, Box, Baseline in emotion recognition across seven different emotional categories.
  }
  \label{ablation_box_number_landmarks_bar}
\end{figure}

Fig.~\ref{total_3_methods} reveals a comparative analysis of the performance across different emotional categories using three different visual prompts for GPT-4V: SoV (Ours), RedCircle, and ReCLIP. In this assessment, SoV outperforms the other methods in most emotional categories. Notably, SoV scores exceptionally well for `Happy' and `Neutral' emotions, suggesting a robust capability in recognizing these emotions accurately. RedCircle and ReCLIP show similar performance for `Fear' and `Sad', but lag significantly behind SoV in `Happy' and `Neutral'. For `Angry', SoV and ReCLIP perform almost equally well, both surpassing RedCircle. This chart underscores SoV's strengths in nuanced emotion recognition, particularly in distinguishing positive and neutral states effectively.

\subsection{Ablation Study}
We investigate the effects of different vision prompts on GPT-4V. The data from Table \ref{ablation_table_vision_prompts} demonstrates how various vision prompts affect an emotion recognition system's performance across different difficulty levels. Baseline prompt shows moderate effectiveness with an accuracy of 48.85\% in Easy, 47.95\% in Medium, but drops to 32.45\% in Hard scenarios. The overall accuracy is 44.44\%, with Recall@1 similarly distributed, suggesting it performs consistently across different complexities but struggles with harder categories. Introduction of bounding boxes slightly reduces accuracy in simpler categories but improves Recall@1, suggesting enhanced precision in pinpointing relevant emotions. Notably, combining box and number prompts markedly boosts performance across all categories, particularly in challenging environments where accuracy rises significantly from 32.45\% to 42.10\% and Recall@1 from 11.36\% to 15.57\%. The SoV prompt outperforms all other methods, achieving peak accuracies of 60.91\% in Easy and 50.00\% in Hard, along with the highest Recall@1 figures, confirming its effectiveness in accurately interpreting emotions even in the most difficult settings.
\begin{figure}[tb]
  \centering
  \includegraphics[width = \linewidth]{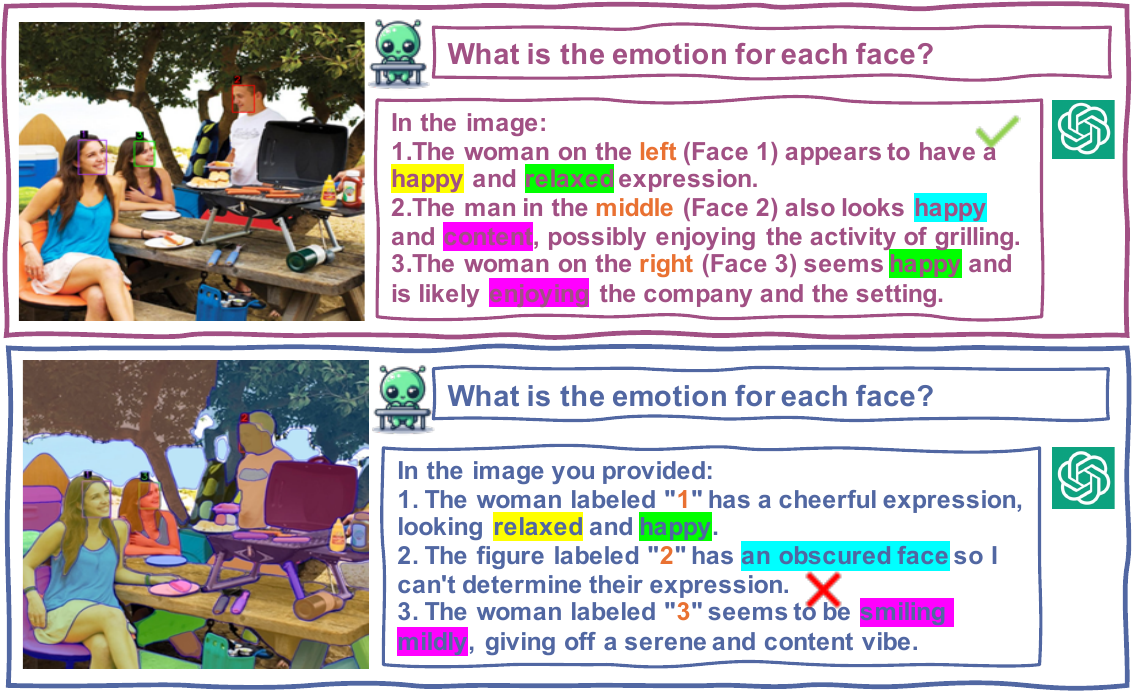}
  \caption{The impacts of segmentation masks for emotion recognition. \textbf{Top:} SoV provides a clearer view for emotion recognition. \textbf{Bottom:} the segmentation masks obscure parts of their faces, making it more challenging to accurately discern these emotions, especially for Person 2. In addition, the
added segmentation masks also result in a lack
of precise context.
  }
  \label{ablation_mask_impacts}
\end{figure}

Fig.~\ref{ablation_box_number_landmarks_bar} showcases an ablation study that compares the effectiveness of different vision prompts on GPT-4V. The GPT-4V+SoV configuration consistently outperforms the other methods in nearly all emotional categories, particularly excelling in `Neutral' and `Angry'. While the Box + Number prompts demonstrates moderate success in `Happy' and `Surprise', it still falls short compared to SoV prompts in other emotion categories. This result highlights that after adding extra facial landmarks, the VLLMs can capture more nuanced and varied emotional states. 

\subsection{Qualitative Observations}
The image in Fig. \ref{ablation_mask_impacts} shows three people at a picnic setting, each displaying happy and relaxed expression, which suggests they are enjoying the outing. However, due to the segmentation masks applied, the masks obscure parts of their faces, making it more challenging to accurately discern these emotions, especially for person 2. In addition, the added segmentation masks also result in a lack of precise context.

\section{Limitations}
We suggest the implementation of Set-of-Vision prompting to bridge visual and textual prompts. However, a challenge arises as it is difficult to precisely describe visual prompts, such as the shape, location, or color of a bounding box and facial landmarks, in language. This issue might require encoding the visual prompts and fine-tuning the entire model for better accuracy. Moreover, this method is computationally intensive, potentially limiting its scalability and practicality in real-time applications, especially when handling large datasets or streaming videos such as tracking one person's emotion in different video frames.

\section{Conclusion}
In conclusion, our Set-of-Vision prompting (SoV) approach significantly advances the field of emotion recognition within VLLMs by addressing critical challenges in spatial localization and global context preservation. By leveraging spatial information such as bounding boxes and facial landmarks, SoV enhances zero-shot emotion recognition accuracy, ensuring precise face count and emotion categorization. Our face overlap handling algorithm and combined text-vision prompting strategy further refine the recognition process, highlighting the efficacy of integrating visual prompts in VLLMs for more accurate and detailed emotion analysis. This approach not only preserves the enriched image context but also offers a solution for detailed and nuanced emotion recognition, underscoring its potential impact on various applications within computer vision and natural language processing.



\bibliography{emnlp_llms}

\cleardoublepage
\appendix
\section{Appendix}
\label{sec:appendix}
\subsection{Dataset details}
In the dataset Table \ref{dataset_introduction}, the "easy" dataset includes pictures with three or fewer faces. The "medium" dataset includes pictures with 3 to 7 faces. The "hard" dataset includes pictures with more than 7 faces. Easy dataset contains 76 images with a total of 174 faces. Medium dataset consists of 34 images featuring 171 faces. Hard dataset is the smallest set, comprising 9 images but still containing a significant number of faces (114). The table categorizes the datasets based on the complexity and density of faces in the images, which likely affects the challenge level for the model's emotion recognition capabilities. The usage of SoV prompts across all categories suggests a consistent testing approach, aiming to evaluate how well the model can interpret and predict emotions without prior specific training on these images (zero-shot learning). The metrics, Accuracy and Recall, are chosen to assess the model's precision in correctly identifying emotions and its ability to retrieve relevant instances across the datasets, respectively.
\label{appendix_dataset_details}
\begin{table}[htbp]
\tiny
\centering
\caption{The table presents the dataset details used for testing a model on the task of zero-shot emotion recognition, structured across three different levels of difficulty: Easy, Medium, and Hard.}
\begin{tabular}{|c|c|c|c|c|c|c|c|c|c|c|c|}
\hline
Dataset & \#Images & \#Faces & Prompts & Metrics\\
\hline
Easy & 76 & 174 & SoV& Accuracy \& Recall\\
Medium & 34 & 171 & SoV & Accuracy \& Recall\\
Hard & 9 & 114 & SoV &  Accuracy \& Recall\\
Total & 119 & 459 & SoV &  Accuracy \& Recall\\
\hline
\end{tabular}
\label{dataset_introduction}
\end{table}
\subsection{Scene based emotion recognition}
\begin{figure*}[tb]
  \centering
  \includegraphics[width = \linewidth]{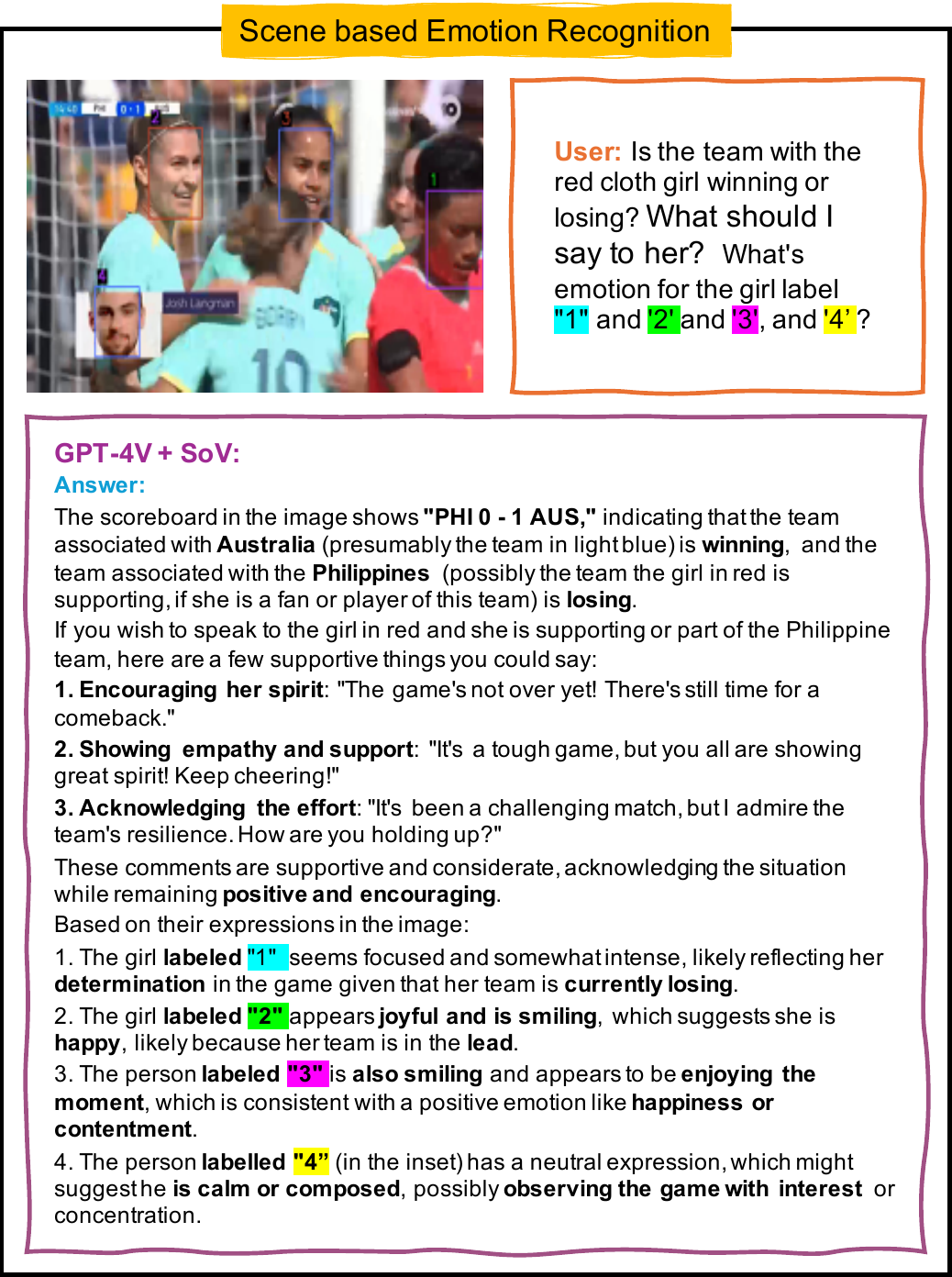}
  \caption{Scene based Emotion Recognition
  }
  \label{scene_based_emotion_recogniton}
\end{figure*}

The image in Fig. \ref{scene_based_emotion_recogniton} captures a moment from a sports event, specifically a match between teams from Australia and the Philippines, with Australia leading based on the scoreboard ("PHI 0 - 1 AUS"). The scene includes various spectators and players, each showing distinct emotions which can aid in understanding the context and overall sentiment related to the ongoing game. Including scene context can significantly enhance the performance of language and image models (LLMs) in recognizing emotions. By understanding not just the facial expressions but also the situational context (such as the score in a sports game), models can make more accurate inferences about the probable emotions being displayed. In addition, environmental cues like scoreboards, team colors, and body language provide additional data points that help in accurately deducing the emotional state of individuals in group settings. Our SoV approach can bridge the gap between purely facial expression-based recognition and a more situation-aware understanding, leading to more nuanced and accurate emotion recognition capabilities in VLLMs.

\begin{figure*}[tb]
  \centering
  \includegraphics[width = \linewidth]{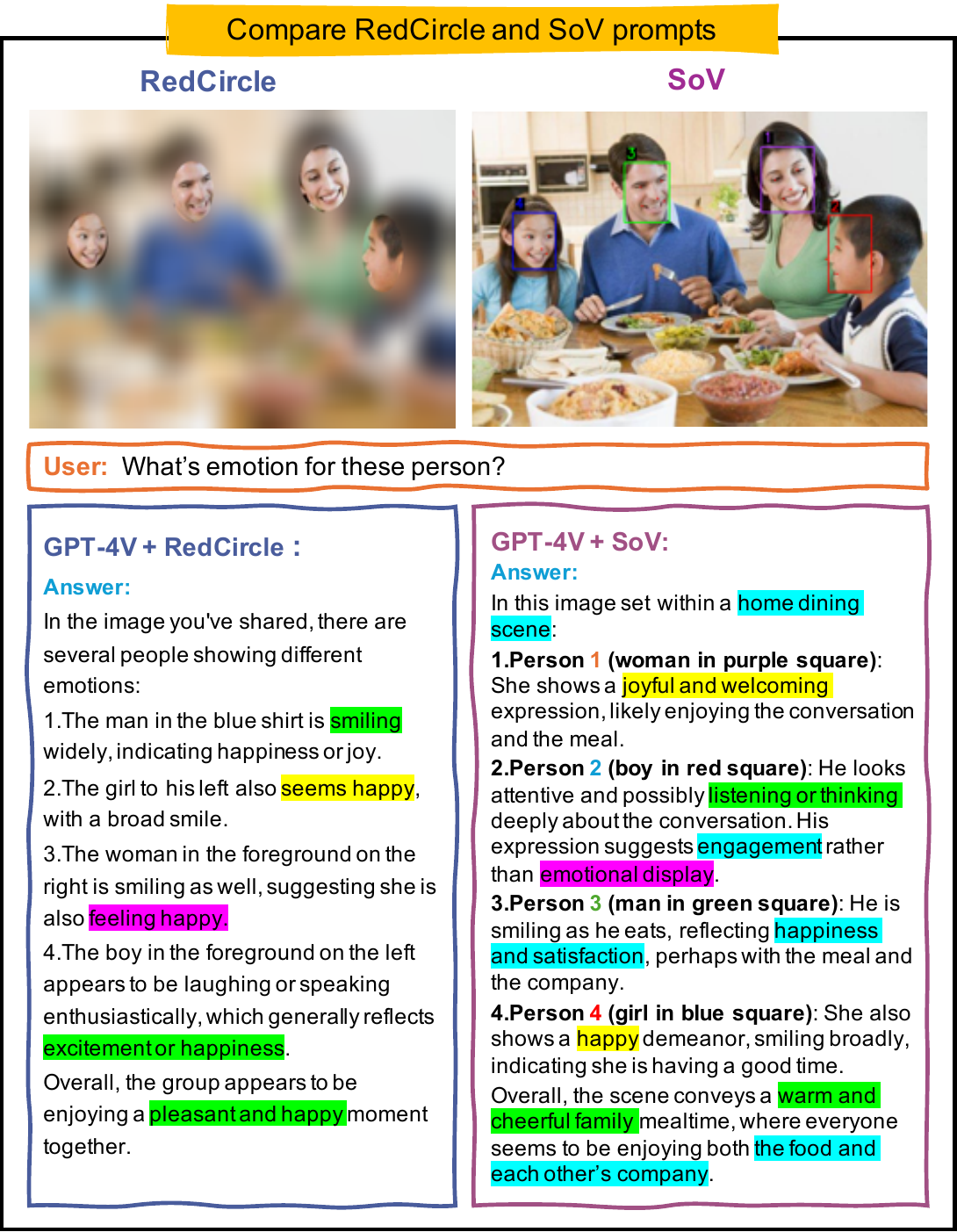}
  \caption{Compare RedCircle and SoV prompts for emotion recognition. 
  }
  \label{redcircle_sov_appendix}
\end{figure*}

In Fig. \ref{redcircle_sov_appendix}, the SoV method, by focusing on specific individuals within the image and retaining the clarity of the background context, offers a comprehensive approach to emotion recognition. This allows for a detailed analysis of not just the visible facial expressions, but also the situational context and interactions among individuals, providing a more nuanced understanding of the emotions being conveyed. For example, in the case of Person 2, despite a partially visible face, the SoV approach leverages additional contextual cues to deduce the emotion more accurately. In contrast, the RedCircle approach primarily relies on clear facial expressions by blurring background details, potentially overlooking subtler emotional cues embedded in the environment and body language.

\begin{figure*}[tb]
  \centering
  \includegraphics[width = \linewidth]{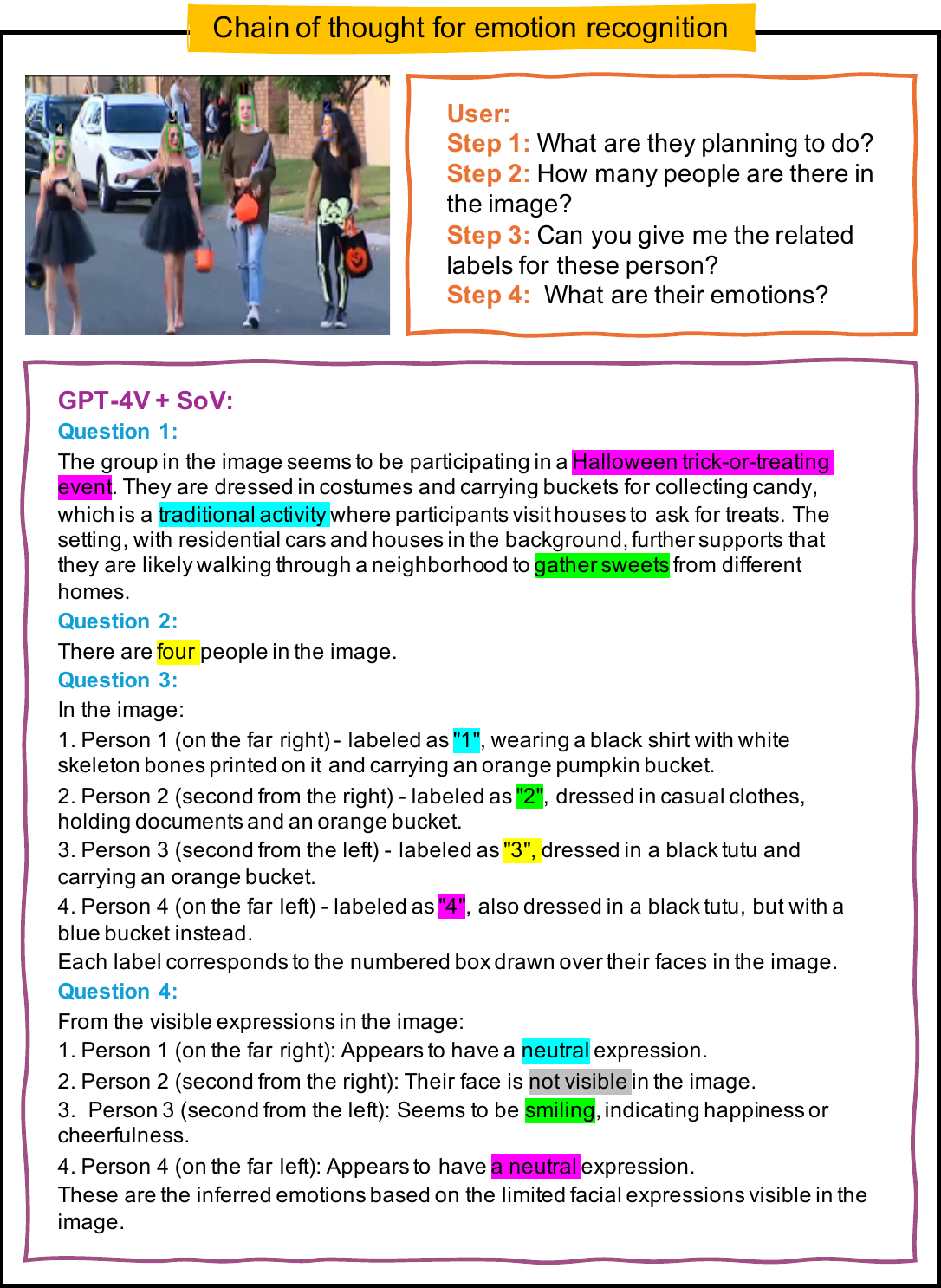}
  \caption{Chain of thought for emotion recognition
  }
  \label{chain_of_thought}
\end{figure*}
\subsection{Chain of thought for emotion recognition}
The "Chain of Thought" approach, as illustrated in Fig. \ref{chain_of_thought}, offers advantages by breaking down the thought process into sequential, logical steps. This method enhances accuracy in interpreting emotions by considering various contextual clues such as activities, and body language, which are then systematically analyzed. For example, the analysis first identifies the activity (Halloween trick-or-treating) and setting, which sets the emotional backdrop. Subsequently, it classifies each individual based on visible attributes and costumes, leading to more nuanced emotion recognition. This step-by-step reasoning mirrors human cognitive processes, allowing for more refined and contextually appropriate interpretations of emotions, such as distinguishing neutral expressions from smiles, even in a complex social setting like Halloween, where expressions might otherwise be ambiguous.
\begin{figure*}[tb]
  \centering
  \includegraphics[width = \linewidth]{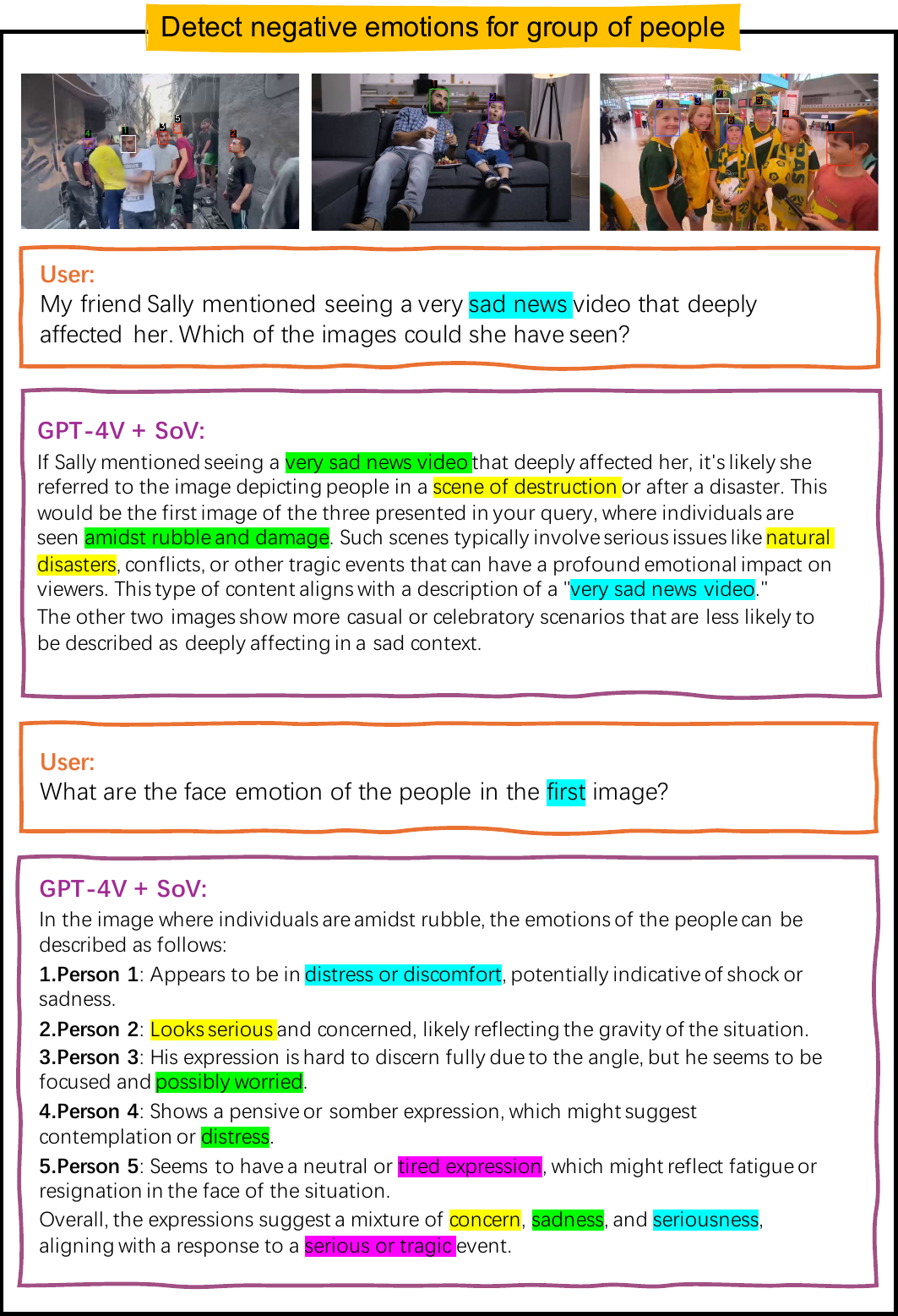}
  \caption{Detect negative emotions for different people
  }
  \label{negative_emotion_recognition}
\end{figure*}
\begin{figure*}[tb]
  \centering
  \includegraphics[width = \linewidth]{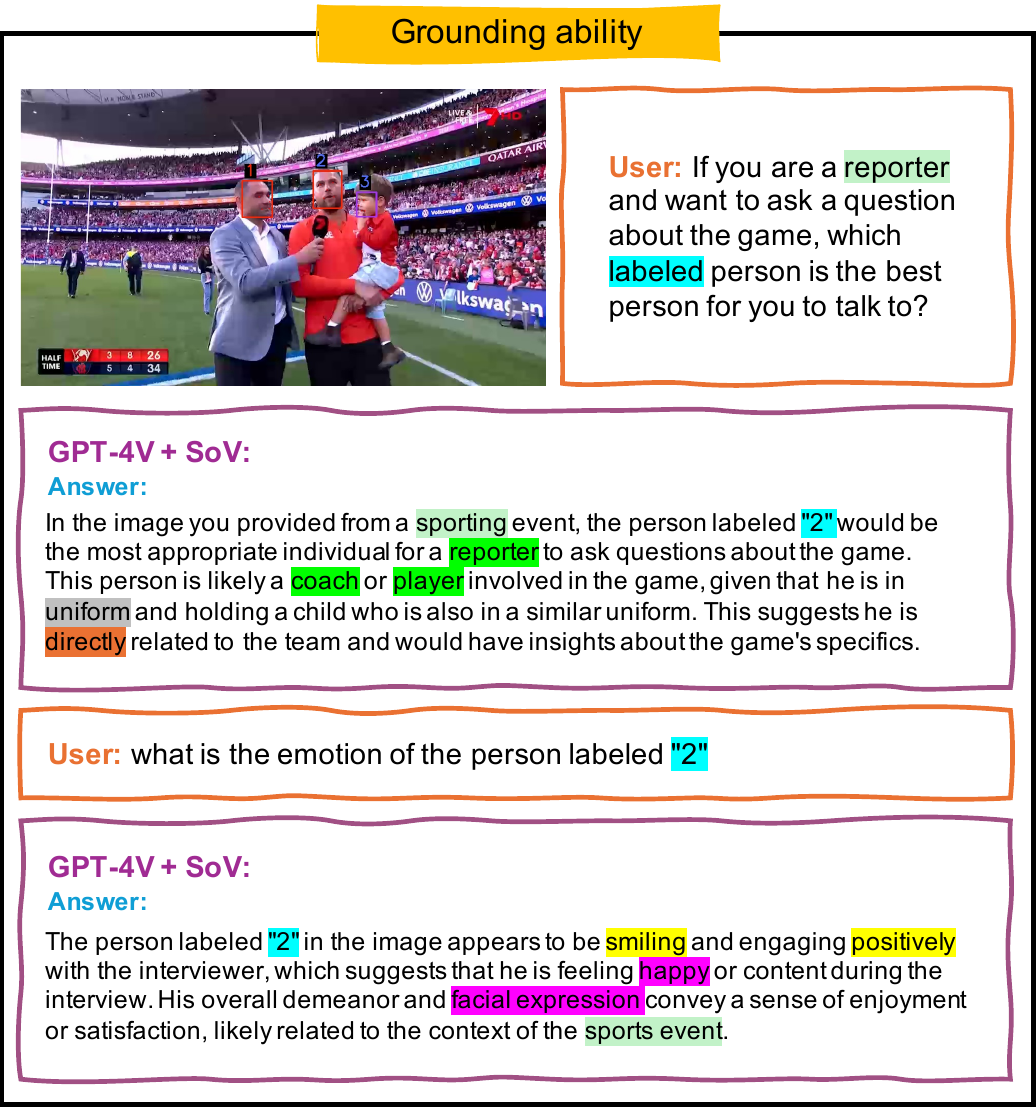}
  \caption{Grounding ability
  }
  \label{grounding_truth_appendix}
\end{figure*}
\end{document}